\documentclass[11pt]{article}

\usepackage[]{acl}

\usepackage{times}
\usepackage{latexsym}

\usepackage[T1]{fontenc}

\usepackage[utf8]{inputenc}

\usepackage{microtype}

\usepackage{enumitem}

\usepackage{booktabs}
\usepackage{url}
\usepackage{times}
\usepackage{latexsym}
\usepackage{hyperref}

\usepackage{amsmath,amssymb}
\usepackage{graphicx}
\usepackage{tabularx}
\usepackage{multirow}
\usepackage{pifont}
\usepackage{soul}

\usepackage[toc,page]{appendix}
\usepackage{xcolor}
\usepackage{floatrow}
\usepackage{makecell}
\usepackage{xspace}
\usepackage{colortbl}

\newfloatcommand{capbtabbox}{table}[][\FBwidth]

\usepackage{arydshln}

\usepackage{minibox}
\usepackage{tabularx,ragged2e}
\usepackage{tabularx} %
\usepackage{algorithm}
\usepackage[noend]{algpseudocode}
\usepackage{bbm}

\DeclareMathOperator*{\argmax}{argmax}
\DeclareMathOperator*{\score}{score}

\DeclareMathOperator*{\E}{\mathbb{E}}
\newcommand{\norm}[1]{\left\lVert#1\right\rVert}
\DeclareMathOperator{\BERT}{\texttt{BERT}}

\DeclareMathOperator{\relaxedbernoulli}{RelaxedBernoulli}
\renewcommand{\vec}[1]{\mathbf{#1}}

\newcommand\ti[1]{\textit{#1}}

\newcommand\tf[1]{\textbf{#1}}
\newcommand\ttt[1]{\texttt{#1}}

\newcommand{\tableindent}{~~}

\newcommand{\cmark}{\textcolor{green}{\ding{51}}\xspace}%
\newcommand{\xmark}{\textcolor{red}{\ding{55}}\xspace}%

\definecolor{aquamarine}{rgb}{0.5, 1.0, 0.83}

\DeclareRobustCommand{\hlaqua}[1]{{\sethlcolor{yellow}\hl{#1}}}

\newcommand{\spectra}{SPECTRA\xspace}
\newcommand{\fc}{FC\xspace}
\newcommand{\fcsup}{FC-sup\xspace}
\newcommand{\vib}{VIB\xspace}
\newcommand{\vibsup}{VIB-sup\xspace}
\newcommand{\addtextadv}{AddText-Adv\xspace}
\newcommand{\addtextwiki}{AddText-Wiki\xspace}
\newcommand{\addtextrand}{AddText-Rand\xspace}

\newcommand{\CC}{\cellcolor{gray!15}}

\title{Can Rationalization Improve Robustness?}

\author{Howard Chen \quad Jacqueline He \quad Karthik Narasimhan \quad Danqi Chen\\
  \large{Department of Computer Science, Princeton University}\\
  \texttt{\{howardchen, karthikn, danqic\}@cs.princeton.edu}\\
  \texttt{jyh@princeton.edu}
}

\begin{document}
\maketitle

\begin{abstract}

A growing line of work has investigated the development of neural NLP models that can produce \ti{rationales}---subsets of input that can explain their model predictions.
In this paper, we ask whether such rationale models can provide robustness to adversarial attacks in addition to their interpretable nature. Since these models need to first generate rationales (``rationalizer'') before making predictions (``predictor''), they have the potential to ignore noise or adversarially added text by simply masking it out of the generated rationale. To this end, we systematically generate various types of `AddText' attacks for both token and sentence-level rationalization tasks and perform an extensive empirical evaluation of state-of-the-art rationale models across five different tasks.
Our experiments reveal that rationale models show promise in improving robustness but struggle in certain scenarios---e.g., when the rationalizer is sensitive to position bias or lexical choices of the attack text. Further, leveraging human rationales as supervision does not always translate to better performance. Our study is a first step towards exploring the interplay between interpretability and robustness in the rationalize-then-predict framework.\footnote{Our code is publicly available at: \url{https://github.com/princeton-nlp/rationale-robustness}.}

\end{abstract}

\begin{figure}[th]
\includegraphics[width=1.12\linewidth, height=5.2cm]{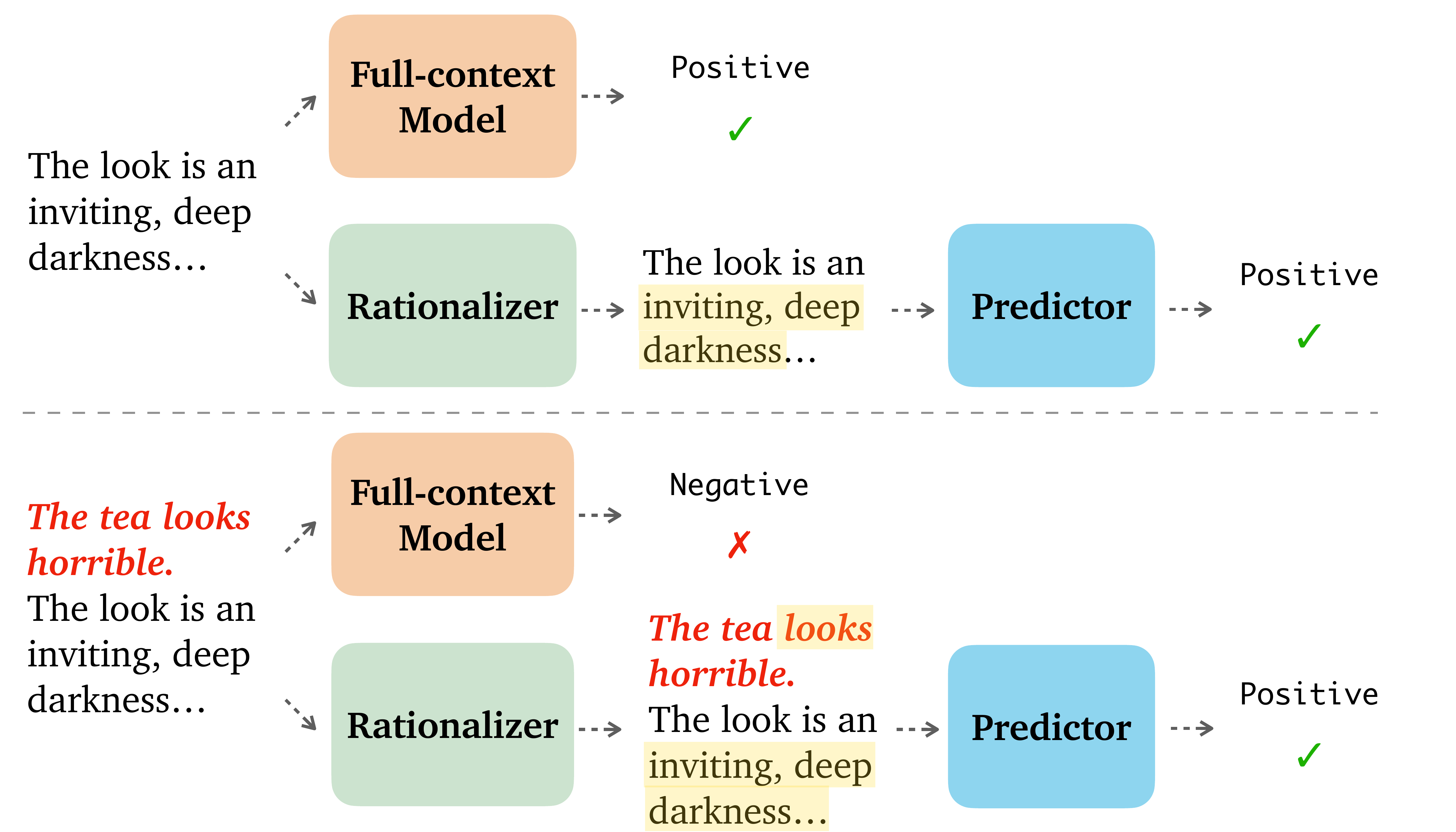}
\caption{Top: input text is processed by a rationale model (rationalizer and predictor) and a full-context model (making predictions based on the whole input) separately in a \ti{beer review} sentiment classification dataset. Both models make correct predictions. Bottom: when an attack sentence ``\ti{The tea looks horrible.}''  is inserted, the full-context model fails. The rationalizer successfully excludes the negative-sentiment word ``horrible'' from the selected rationales (yellow highlights). The predictor is hence not distracted by the attack sentence.
}
\label{fig:teaser}
\end{figure}

\section{Introduction}
\label{sec:intro}

Rationale models aim to introduce a degree of interpretability into neural networks by implicitly baking in explanations for their decisions~\cite{lei2016rationale,bastings2019interpretable,jain2020learning}.
These models are carried out in a two-stage `rationalize-then-predict' framework, where the model first selects a subset of the input as a \emph{rationale} and then makes its final prediction for the task solely using the rationale. A human can then inspect the selected rationale to verify the model's reasoning over the most relevant parts of the input for the prediction at hand.

While previous work has mostly focused on the plausibility of extracted rationales and whether they represent faithful explanations~\cite{deyoung2020eraser}, we ask the question of how rationale models behave under adversarial attacks (i.e., do they still provide plausible rationales?) and whether they can help improve robustness (i.e., do they provide better task performance?). Our motivation is that the two-stage decision-making could help models ignore noisy or adversarially added text within the input. For example, Figure~\ref{fig:teaser} shows a state-of-the-art rationale model~\cite{paranjape2020ibrationale} smoothly handles input with adversarially added text by selectively masking it out during the rationalization step. Factorizing the rationale prediction from the task itself effectively `shields' the predictor from having to deal with adversarial inputs.

To answer these questions, we first generate adversarial tests for a variety of popular NLP tasks (\S\ref{sec:robust_test}). We focus specifically on model-independent, `AddText' attacks~\cite{jia2017advexampleqa}, which augment input instances with noisy or adversarial text at \ti{test time}, and study how the attacks affect rationale models both in their prediction of rationales and final answers. For diversity, we consider inserting the attack sentence at different positions of context, as well as three types of attacks: random sequences of words, arbitrary sentences from Wikipedia, and adversarially-crafted sentences.

We then perform an extensive empirical evaluation of multiple state-of-the-art rationale models~\cite{paranjape2020ibrationale,guerreiro2021spectra}, across five different tasks that span review classification, fact verification, and question answering (\S\ref{sec:models_datasets}). In addition to the attack's impact on task performance, we also assess rationale prediction by defining metrics on gold rationale coverage and attack capture rate.
We then investigate the effect of incorporating human rationales as supervision, the importance of attack positions, and the lexical choices of attack text. Finally, we investigate an idea of improving the rationalizer by adding augmented pseudo-rationales during training (\S\ref{sec:art}).

Our key findings are the following:
\begin{enumerate}[nosep]
	\item Rationale models show promise in providing robustness. Under our strongest type of attack, rationale models in many cases achieve less than 10\% drop in task performance while full-context models suffer more (11\%--27\%).
	\item However, robustness of rationale models can vary considerably with the choice of lexical inputs for the attack and is quite sensitive to the attack position.
	\item Training models with explicit rationale supervision does not guarantee better robustness to attacks. In fact, their accuracy drops under attack are higher by 4-10 points compared to rationale models without supervision.
	\item Performance under attacks is  significantly improved if the rationalizer can effectively mask out the attack text. Hence, our simple augmented-rationale training strategy can effectively improve robustness (up to 4.9\%).
\end{enumerate}

Overall, our results indicate that while there is promise in leveraging rationale models to improve robustness, current models may not be sufficiently equipped to do so. Furthermore, adversarial tests may provide an alternative form to evaluate rationale models in addition to prevalent plausability metrics that measure agreement with human rationales. We hope our findings can inform the development of better methods for rationale predictions and instigate more research into the interplay between interpretability and robustness.

\section{Related Work}
\label{sec:related}

\paragraph{Rationalization} There has been a surge of work on explaining predictions of neural NLP systems, from post-hoc explanation methods~\cite{ribeiro2016should,alvarez2017causal}, to analysis of attention mechanisms~\cite{jain2019attention,serrano2019attention}. We focus on \ti{extractive rationalization}~\cite{lei2016rationale}, which generates a subset of inputs or highlights as ``rationales'' such that the model can condition predictions on them.
Recent development has been focusing on improving joint training of rationalizer and  predictor components~\cite{bastings2019interpretable,yu2019rethinking,jain2020learning,paranjape2020ibrationale,guerreiro2021spectra,sha2021learnfrombest}, or extensions to text matching~\cite{swanson2020rationalizing} and sequence generation~\cite{vafa2021rationales}. These rationale models are mainly compared based on predictive performance, as well as agreement with human annotations~\cite{deyoung2020eraser}. In this work, we question how rationale models behave under adversarial attacks and whether they can provide robustness benefits through rationalization.

\paragraph{Adversarial examples in NLP} Adversarial examples have been designed to reveal the brittleness of state-of-the-art NLP models. A flood of research has been proposed to generate different adversarial attacks~\cite[][\emph{inter alia}]{jia2017advexampleqa,iyyer2018adversarial,belinkov2018synthetic,ebrahimi2018hotflip}, which can be broadly categorized by types of input perturbations (sentence-, word- or character-level attacks), and access of model information (black-box or white-box). In this work, we focus on \ti{model-independent}, label-preserving attacks, in which we \ti{insert} a random or an adversarially-crafted sentence into input examples~\cite{jia2017advexampleqa}. We hypothesize that a good extractive rationale model is expected to learn to ignore these distractor sentences and hence achieve better performance under attacks.

\paragraph{Interpretability and robustness} A key motivation of our work is to bridge the connection between interpretability and robustness, which we believe is an important and under-explored area. \newcite{alvarez2018robustness} argue that robustness of explanations is a key desideratum for interpretability. \newcite{slack2020fooling} explore unreliability of attribution methods against input perturbations. \newcite{camburu2020makeup} introduce an adversarial framework to sanity check models against their generated inconsistent free-text explanations. \newcite{zhou2020dofeatureattn} propose to evaluate attribution methods through dataset modification.
\newcite{noack2021empirical} show that image recognition models can achieve better adversarial robustness when they are trained to have  interpretable gradients.
To the best of our knowledge, we are the first to quantify the performance of rationale models under textual adversarial attacks and understand whether rationalization can inherently provide robustness.

\section{Background}
\label{sec:background}
Extractive rationale models\footnote{Abstractive models~\cite{wiegreffe2021association,narang2020wt5}, which generate rationales as free text, are an alternative class of models that we do not consider in this work.} output predictions through a two-stage process: the first stage (``rationalizer'') selects a subset of the input as a \textit{rationale}, while the second stage (``predictor'') produces the prediction using only the rationale as input. \textit{Rationales} can be any subset of the input, and we characterize them roughly into either token-level or sentence-level rationales, which we will both investigate in this work. The task of predicting rationales is often framed as a binary classification problem over each atomic unit depending on the type of rationales. The rationalizer and the predictor are often trained jointly using task supervision, with gradients back-propagated through both stages. We can also provide explicit rationale supervision, if human annotations are available.

\subsection{Formulation}
Formally, let us assume a supervised classification dataset $\mathcal{D} = \{(x, y)\}$
, where each input $x = x_1, x_2, ..., x_T$ is a concatenation of $T$ sentences and each sentence $x_t = (x_{t,1}, x_{t,2}, ... x_{t,n_t})$ contains $n_t$ tokens, and $y$ refers to the task label. A rationale model consists of two main components: 1) a rationalizer module $z = R(x; \theta)$, which generates a discrete mask $z \in \{0, 1\}^L$ such that $z \odot x$ selects a subset from the input ($L = T$ for sentence-level rationalization or $L = \sum_{i=1}^{T}{n_i}$ for token-level rationales), and 2) a predictor module $\hat{y} = C(x, z; \phi)$ that makes a prediction $\hat{y}$ using the generated rationale $z$. The entire model $M(x) = C(R(x)) $ is trained end-to-end using the standard cross-entropy loss. We describe detailed training objectives in \S\ref{sec:models_datasets}.

\subsection{Evaluation}
Rationale models are traditionally evaluated along two dimensions: a) their downstream task performance, and b) the quality of generated rationales.  To evaluate rationale quality, prior work has used metrics like token-level F1 or Intersection Over Union (IOU) scores between the predicted rationale and a human rationale \citep{deyoung2020eraser}:
\begin{equation}
\text{IOU} = \frac{\lvert z \cap z^* \rvert}{\lvert z \cup z^* \rvert} \notag,
\end{equation}
where $z^*$ is the human-annotated gold rationales.

A good rationale model should not sacrifice task performance while generating rationales that concur with human rationales. However, metrics like F1 score may not be the most appropriate way to capture this as it only captures \textit{plausibility} instead of \ti{faithfulness} \citep{jacovi2020faithful}.

\section{Robustness Tests for Rationale Models}
\label{sec:robust_test}

\subsection{AddText Attacks}

Our goal is to construct attacks that can test the capability of extractive rationale models to ignore spurious parts of the input.
Broadly, we used two guiding criteria for selecting the type of attacks:
    1) they should be additive since an extractive rationale model can only ``ignore'' the irrelevant context. For other attacks such as counterfactually edited data (CAD)~\cite{kaushik2020cad}, even if the rationalizer could identify the edited context, the predictor is not necessarily strong enough to reason about the counterfactual text;
    2) they should be model-independent since our goal is to compare the performance across different types of rationale and baseline models. Choosing strong gradient-based attacks~\cite{ebrahimi2018hotflip, wallace2019advtrigger} would probably break all models, but that is beyond the scope of our hypothesis. An attack is suitable as long as it reduces performance of standard classification models by a non-trivial amount (our attacks reduce performance by 10\%--36\% absolute).

Keeping these requirements in mind, we focus on label-preserving text addition attacks~\citet{jia2017advexampleqa}, which can test whether rationale models are invariant to the addition of extraneous information and remain consistent with their predictions.
Attacks are only added at test time and are not available during model training.

\paragraph{Attack construction}

Formally, an AddText attack $A(x)$ modifies the input $x$ by adding an attack sentence $x_{\text{adv}}$, without changing the ground truth label $y$. In other words, we create new perturbed test instances $(A(x),y)$ for the model to be evaluated on.
While some prior work has considered the addition of a few tokens to the input~\cite{wallace2019advtrigger}, we add complete sentences to each input, similar to the attacks in \citet{jia2017advexampleqa}. This prevents unnatural modifications to the existing sentences in the original input $x$ and also allows us to test both token-level and sentence-level rationale models (\S\ref{subsec:models}). We experiment with adding the attack sentence $x_{\text{adv}}$ either at the beginning or the end of the input $x$.\footnote{In \S\ref{sec:sensitivity_attack_pos}, we also consider inserting the attack sentence at a random position for studying the effect of attack positions.}

\paragraph{Types of attacks}
We explore three different types of attacks: (1) \textbf{\addtextrand}: we simply add a random sequence of tokens uniformly sampled from the task vocabulary. This is a weak attack that is easy for humans to spot and ignore since it does not guarantee grammaticality or fluency.
(2) \textbf{\addtextwiki}: we add an arbitrarily sampled sentence from English Wikipedia into the task input (e.g., ``Sonic the Hedgehog, designed for \ldots''). This attack is more grammatical than AddText-Rand, but still adds text that is likely irrelevant in the context of the input $x$.
(3) \textbf{\addtextadv}: we add an adversarially constructed sentence that has significant lexical overlap with tokens in the input $x$ while ensuring the output label is unchanged. This type of attack is inspired by prior attacks such as AddOneSent~\cite{jia2017advexampleqa} and is the strongest attack we consider since it is more grammatical, fluent, and contextually relevant to the task.
    The construction of this attack is also specific to each task we consider, hence we provide examples listed in Table~\ref{table:task_attack} and more details in \S\ref{subsec:tasks}.

\subsection{Robustness Evaluation}
\label{sec:robust_eval}
We measure the robustness of rationale models under our attacks along two dimensions: \ti{task performance}, and \ti{generated rationales}. The change in task performance is simply computed as the difference between the average scores of the model on the original vs perturbed test sets:
\begin{equation}
\begin{aligned}
    \Delta = \frac{1}{|\mathcal{D}|}\sum_{(x, y) \in \mathcal{D}}{f(M(x), y) - f(M(A(x)), y)} \notag,
\end{aligned}
\end{equation}
where $f$ denotes a scoring function (F1 scores in extractive question answering and $\mathbb{I}(y = \hat{y})$ in text classification).
To measure the effect of the attacks on rationale generation, we use two metrics:
\paragraph{Gold rationale F1 (GR)} This is defined as the F1 score between the predicted rationale and a human-annotated rationale, either computed at the token or sentence level. The token-level GR score is equivalent to F1 scores reported in previous work \citep{lei2016rationale, deyoung2020eraser}. A good rationalizer should generate plausible rationales and be not affected by the addition of attack text.

\paragraph{Attack capture rate (AR)} We define AR as the recall of the inserted attack text in the rationale generated by the model:
    \begin{equation}
        \mathrm{AR} = \frac{1}{\lvert \mathcal{D} \rvert} \sum_{(x, y) \sim \mathcal{D}} \frac{\lvert x_{\text{adv}} \cap (z \odot A(x)) \rvert}{\lvert x_{\text{adv}} \rvert} \notag,
    \end{equation}
    where $x_{\text{adv}}$ is the attack sentence added to each instance (i.e., $A(x)$ is the result of inserting $x_{\text{adv}}$ into $x$), $z \odot A(x)$ is the predicted rationale. The metric above applies on both token or sentence level ($\lvert x_{\text{adv}} \rvert = 1$ for sentence-level rationalization and number of tokens in the attack sentence for token-level rationalization).
    This metric allows us to measure how often a rationale model can \textit{ignore} the added attack text---a maximally robust rationale model should have an AR of 0.

\begin{table*}[th]
\centering
\resizebox{1.0\columnwidth}{!}{%
\begin{tabular}{llll}
\toprule
\tf{Dataset} & \tf{Query $\rightarrow$ Attack} & \tf{Full Attacked Input} & \tf{Label} \\
\midrule
\multirow{4}{*}{FEVER} & & \textit{Query}: \textcolor{blue}{Jennifer Lopez was married.} & \multirow{4}{*}{Supports}\\
                       & \textcolor{blue}{Jennifer Lopez was married.} & \textit{Context}: Jennifer Lynn Lopez (born July 24 , 1969), also known \\
                       & $\rightarrow$ \textcolor{red}{Jason Bourne was unmarried.} & as JLo, is an American singer $\ldots$ \sethlcolor{yellow}\hl{She subsequently married} &\\
                       & & \sethlcolor{yellow}\hl{longtime friend Marc Anthony} $\ldots$ \textcolor{red}{Jason Bourne was unmarried.} & \\
\midrule
\multirow{4}{*}{SQuAD} & & \textit{Query}: \textcolor{blue}{Where did Super Bowl 50 take place?} & \multirow{4}{*}{Levi's Stadium}\\
                       & \textcolor{blue}{Where did Super Bowl 50 take place?} & \textit{Context}: Super Bowl 50 was an American football game to\\
                       & $\rightarrow$ \textcolor{red}{The Champ Bowl 40 took place in Chicago.} & determine the champion $\ldots$ was played on February 7, 2016, & \\
                       & & at \sethlcolor{yellow}\hl{Levi's Stadium} $\ldots$ \textcolor{red}{The Champ Bowl} \textcolor{red}{40 took place in Chicago.} & \\
\midrule
\multirow{2}{*}{Beer} & N/A & This beer poured a \sethlcolor{yellow}\hl{very appealing} copper reddish color---it & \multirow{2}{*}{Positive}\\
                      & $\rightarrow$ \textcolor{red}{The tea looks horrible.} & was \sethlcolor{yellow}\hl{very clear} with an average head $\ldots$ \textcolor{red}{The tea looks horrible.}\\
\bottomrule
\end{tabular}}
\caption{AddText-Adv attack applied to three datasets. The query ({\color{blue}blue}) is transformed into an attack ({\color{red}red}). The query together with the context forms the input. The attack is inserted to the context. We only show insertion at the end, but the attack can be inserted at any position between sentences. A model needs to associate the query and the evidence (\sethlcolor{yellow}\hl{ground truth rationale}) in the context
and not be distracted by the inserted attack to make the correct prediction. Note that the Beer dataset doesn't have a query and the attack sentence is dependent on the label (\S\ref{subsec:tasks}).}
\label{table:task_attack}
\end{table*}

\section{Models and Tasks}
\label{sec:models_datasets}

We investigate two different state-of-the-art selective rationalization approaches: 1) sampling-based stochastic binary mask generation \citep{bastings2019interpretable, paranjape2020ibrationale}, and 2) deterministic sparse attention through constrained inference \citep{guerreiro2021spectra}.
We adapt these models, using two separate BERT encoders for the rationalizer and the predictor, and consider training scenarios with and without explicit rationale supervision. We also consider a full-context model as baseline. We provide a brief overview of each model here and leave details including loss functions and training to \S\ref{sec:model_detail}.

\subsection{Models without Rationale Supervision}
\label{subsec:models}

\paragraph{Variational information bottleneck (\vib)}
This model~\cite{paranjape2020ibrationale} imposes a discrete bottleneck objective~\cite{alemi2017deepvib} to select a mask $z \in \{0, 1\}^L$ from the input $x$.
The rationalizer samples $z$ using Gumbel-Softmax and the predictor uses only $z \odot x$ for the final prediction. During inference, we select the top-$k$ scored rationales, where $k$ is determined by the sparsity $\pi$.

\paragraph{Sparse structured text rationalization (\spectra)} This model~\citep{guerreiro2021spectra} extracts a deterministic structured mask $z$ by solving a constrained inference problem by applying factors to the global scoring function while optimizing the end task performance. The entire computation is deterministic and allows for back-propagation through the LP-SparseMAP solver \citep{niculae2020lpsparsemap}. We use the \texttt{BUDGET} factor to control the sparsity $\pi$.

\paragraph{Full-context model (\fc)}
As a baseline, we also consider a full-context model, which makes predictions directly conditioned on the entire input. The model is a standard BERT model which adds task-specific classifiers on top of the encoder~\cite{devlin2019bert}.  The model is trained with a cross-entropy loss using task supervision.

\subsection{Models with Rationale Supervision}
\label{sec:sup_models}
\paragraph{VIB with human rationales (\vibsup)}
When human-annotated rationales $z^*$ are available, they can be used to guide the prediction of the sampled masks $z$ by adding a cross entropy loss between them (more details in \S\ref{sec:model_detail}).
\vibsup leverages this supervision signal to guide rationale prediction.

\paragraph{Full-context model with human rationales (\fcsup)}
We also extend the \fc model to leverage human-annotated rationales supervision during training by adding a linear layer on top of the sentence/token representations. Essentially, it is multi-task learning of rationale prediction and the original task, shared with the same BERT encoder.
The supervision is added by calculating the cross entropy loss between the human-annotated rationales and the predicted rationales (\S\ref{sec:model_detail}).

\subsection{Tasks}\label{subsec:tasks}

We evaluate the models on five datasets that cover both sentence-level (FEVER, MultiRC, SQuAD) and token-level (Beer, Hotel) rationalization (examples in Table~\ref{table:task_attack}).
We summarize the dataset characteristics in Table~\ref{table:dataset_attributes}.
\begin{table}[!t]
\centering
\resizebox{1.0\columnwidth}{!}{%
\begin{tabular}{lcc}
\toprule
    \multirow{2}{*}{Dataset} & Rationale Granularity & \multirow{2}{*}{Task}\\
    & (w/ Human Rationale) & \\
\midrule
    FEVER & Sentence (\cmark) & Fact verification$\dagger$ \\
    MultiRC & Sentence (\cmark) & Question answering$\dagger$ \\
    SQuAD & Sentence (\cmark) & Question answering$\ddagger$ \\
    Beer & Token (\xmark) & Sentiment $\dagger$ \\
    Hotel & Token (\xmark) & Sentiment $\dagger$ \\
\bottomrule
\end{tabular}}
\caption{Dataset characteristics for the five datasets. $\dagger$: classification, $\ddagger$: span prediction tasks.
}
\label{table:dataset_attributes}
\end{table}
\paragraph{FEVER} FEVER is a sentence-level binary classification fact verification dataset from the ERASER benchmark \cite{deyoung2020eraser}.
The input contains a claim specifying a fact to verify and a passage of multiple sentences supporting or refuting the claim. For the \addtextadv attacks, we add modified query text to the claims by replacing nouns and adjectives in the sentence with antonyms from WordNet~\citep{fellbaum1998wordnet}.

\paragraph{MultiRC} MultiRC~\citep{khashabi2018multirc} is a sentence-level multi-choice question answering task (reformulated as `yes/no' questions).
For the \addtextadv attacks, we transform the question and the answer separately using the same procedure we used for FEVER.

\paragraph{SQuAD} SQuAD \citep{rajpurkar2016squad} is a popular question answering dataset. We use the AddOneSent attacks proposed in Adversarial SQuAD \citep{jia2017advexampleqa}, except that they always insert the sentence at the end of the paragraph and we consider inserting at the beginning, the end, and a random position.
Since SQuAD does not contain human rationales, we use the sentence that contains the correct answer span as the ground truth rationale sentence. We report F1 score for SQuAD.

\paragraph{Beer} BeerAdvocate is a multi-aspect sentiment analysis dataset \citep{mcauley2012beer}, modeled as a token-level rationalization task.
We use the \textit{appearance} aspect in out experiments. We convert the scores into the binary labels following~\citet{chang2020invrat}.
This task does not have a query as in the previous tasks, we insert a sentence with the template ``\texttt{\{SUBJECT\}} is \texttt{\{ADJ\}}'' into a negative review where the adjective is positive (e.g., ``The tea looks fabulous.'') and vice versa.
We consider one object ``car'' and eight adjectives (e.g., ``clean/filthy'', ``new/old'').

\paragraph{Hotel} TripAdvisor Hotel Review is also a multi-aspect sentiment analysis dataset \citep{wang2010hotel}. We use the \textit{cleanliness} aspect in our experiments. We generate \addtextadv attacks in the same way as we did for the Beer dataset.
We consider three objects ranging from more relevant words such as ``tea'' to less related word ``carpet'' and six adjectives (e.g., ``pretty/disgusting'', ``good/bad'', ``beautiful/ugly'').

\section{Results}
\label{sec:results}

\begin{figure*}[!h]
  \centering
  \includegraphics[width=1.0\textwidth]{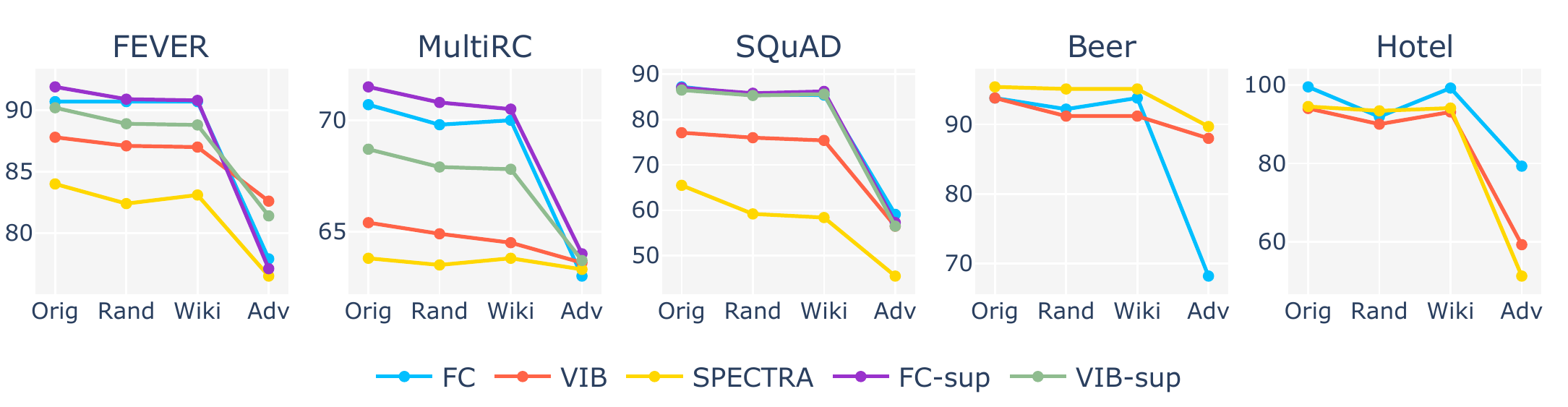}
  \caption{Original performance (Orig) and the three type of attacks  \addtextrand (Rand),  \addtextwiki (Wiki), and \addtextadv (Adv) evaluated on five datasets and all of the models. FC-sup and VIB-sup models used human rationales during training (\S\ref{sec:sup_models}).}
  \label{fig:rand_wiki_adv}
\end{figure*}

\begin{table*}[!h]
\centering
\resizebox{1.0\columnwidth}{!}{%
\begin{tabular}{lccccccccccccccc}
\toprule
    & \multicolumn{3}{c}{FEVER} & \multicolumn{3}{c}{MultiRC} & \multicolumn{3}{c}{SQuAD} & \multicolumn{3}{c}{Beer} & \multicolumn{3}{c}{Hotel} \\
    &Ori & Att & $\Delta \downarrow$ &Ori & Att & $\Delta \downarrow$ &Ori & Att & $\Delta \downarrow$ &Ori & Att & $\Delta \downarrow$ &Ori & Att & $\Delta \downarrow$ \\
\midrule
    Majority & 50.7 & - & - & 54.8 & - & - & - & - & - & 68.9 & - & - & 50.0 & - & -\\
\midrule
    \fc & 90.7 & 77.9 & 12.8 & 70.7 & 63.0 & 7.7 & 87.2 & 59.1 & 28.1 & 93.8 & 59.5 & 34.3 & 99.5 & 79.3 & \bf{20.2}\\
    \vib & 87.8 & 82.6 & \bf{5.2} & 65.4 & 63.6 & 1.8 & 77.1 & 56.5 & 20.6 & 93.8 & 88.0 & 5.8 & 94.0 & 59.3 & 34.8\\
    \spectra & 84.0 & 76.5 & 7.6 & 63.8 & 63.3 & \bf{0.5} & 65.5 & 45.5 & \bf{20.0} & 95.4 & 89.7 & \bf{5.7} & 94.5 & 51.3 & 43.2\\
\midrule
    \fcsup & 91.9 & 77.1 & 14.8 & 71.5 & 64.0 & 7.5 & 87.0 & 57.3 & \tf{29.7} & - & - & - & - & - & -\\
    \vibsup & 90.2 & 81.4 & \tf{8.8} & 68.7 & 63.7 & \tf{5.0} & 86.5 & 56.5 & 30.0 & - & - & - & - & - & -\\
\bottomrule
\end{tabular}}
\caption{Original (Ori) versus attacked (Att) task performance on the five selected datasets under the \addtextadv attack. We report accuracy for all datasets except for SQuAD, which we report F1. The attacked performance is the average of inserting the attack at the start and at the end of the text input.
}
\label{table:main}
\end{table*}

For all attacked test sets, we report the average scores with attack sentence inserted at the beginning and the end of the inputs.
Our findings shed light on the relationship between GR, AR, and drop in performance, which eventually lead to a promising direction to improve performance of rationale models under attacks (\S\ref{sec:art}).

\subsection{Task Performance}\label{sec:overall_results}
Figure~\ref{fig:rand_wiki_adv} summarizes the average scores on all datasets for each model under the three attacks we consider. We first observe that all models (including the full-context models \fc and \fcsup) are mildly affected by \addtextrand and \addtextwiki, with score drops of around $1$-$2\%$. However, the \addtextadv attack leads to more significant drops in performance for all models, as high as $46\%$ for \spectra on the Hotel dataset. We break out the \addtextadv results in a more fine-grained manner in Table~\ref{table:main}. Our main observation is that the rationale models (\vib, \spectra, \vibsup) are generally more robust than their non-rationale counterparts (\fc, \fcsup) on four out of the five tasks, and in some cases dramatically better. For instance, on the Beer dataset, \spectra only suffers a 5.7\% drop (95.4 $\rightarrow$ 89.7) compared to \fc's huge 34.3\% drop (93.8 $\rightarrow$ 59.5) under attack. The only exception is the Hotel dataset, where both the \vib and \spectra models perform worse under attack compared to \fc. We analyze this phenomena and provide a potential reason below.

\subsection{Robustness Evaluation: GR vs AR}\label{sec:gr_ar_tradeoff}

\begin{table*}[!t]
\centering
\resizebox{1.0\columnwidth}{!}{%
\begin{tabular}{lcccccccccc}
\toprule
    & \multicolumn{2}{c}{{FEVER}} & \multicolumn{2}{c}{{MultiRC}} & \multicolumn{2}{c}{{SQuAD}} & \multicolumn{2}{c}{{Beer}} & \multicolumn{2}{c}{{Hotel}}\\
    & {GR $\uparrow$} & {AR $\downarrow$} & {GR $\uparrow$} & {AR $\downarrow$} & {GR $\uparrow$} & {AR $\downarrow$} & {GR $\uparrow$} & {AR $\downarrow$} & {GR $\uparrow$} & {AR $\downarrow$}\\
\midrule
    \vib     & 36.9$\rightarrow$30.3 & 59.4 & 15.8$\rightarrow$13.9 & 35.8 & 86.2$\rightarrow$84.9 & 63.7 & 20.5$\rightarrow$18.1 & 11.9 & 23.5$\rightarrow$22.6 & 18.4\\
    \spectra & 26.9$\rightarrow$21.5 & 40.6 & 11.9$\rightarrow$11.8 & 22.6 & 67.1$\rightarrow$60.8 & 52.6 & 28.6$\rightarrow$27.8 & 15.2 & 19.5$\rightarrow$18.3 & 31.6 \\
\midrule
   \fcsup & 51.5$\rightarrow$45.5 & 65.9 & 50.0$\rightarrow$42.7 & 55.7 & 99.6$\rightarrow$98.8 & 97.8 & - & - & - & -\\
   \vibsup & 50.6$\rightarrow$44.3 & 67.0 & 36.1$\rightarrow$22.7 & 58.7 & 99.5$\rightarrow$97.8 & 97.2 & - & - & - & - \\
\bottomrule
\end{tabular}}
\caption{Gold rationale F1 (GR) (original$\rightarrow$perturbed input) and attack capture rate (AR) for the AddText-Adv attack on the five tasks (defined in \S\ref{sec:robust_eval}). The reported number is the average of inserting the attack at the start and at the end of the text input.}
\label{table:capture}
\end{table*}

In Table~\ref{table:capture}, we report the Gold Rationale F1 (GR) and Attack Capture Rate (AR) for all models.
When attacks are added, GR consistently decreases for all tasks. However, AR ranges widely across datasets. VIB and SPECTRA have lower AR and higher GR compared to FC-sup across all tasks, which is correlated with their superior robustness to AddText-Adv attacks.

Next, we investigate the poor performance of \vib and \spectra on the Hotel dataset by analyzing the choice of words in the attack. %
Using the template ``My car is \ttt{\{ADJ\}}.'', we measure the percentage of times the rationalizer module selects the adjective as part of its rationale. When the adjectives are ``dirty'' and ``clean'', the \vib model selects them a massive $98.5\%$ of the time. For ``old'' and ``new'', \vib still selects them $50\%$ of the time.
On the other hand, the \vib model trained on Beer reviews with attack template ``The tea is \ttt{\{ADJ\}}.'' only selects the adjectives $20.5\%$ of the time (when the adjectives are ``horrible'' and ``fabulous''). This shows that the bad performance of the rationale models on the Hotel dataset is due to their inability to ignore task-related adjectives in the attack text, hinting that the lexical choices made in constructing the attack can largely impact robustness.

\begin{table}[!t]
\centering
\resizebox{0.9\columnwidth}{!}{%
\begin{tabular}{lcc}
\toprule
    & \multicolumn{1}{c}{\vib} & \multicolumn{1}{c}{\vibsup} \\
    & Acc (\%) & Acc (\%) \\
\midrule
    Original & 87.8  & 90.2 \\
\midrule
    Overall Attack & 83.0 (100\%) & 84.9 (100\%)\\
    \tableindent G \cmark~~A \cmark & 83.3 ~~(34\%) & 85.5 ~~(77\%) \\
    \tableindent G \cmark~~A \xmark & \hlaqua{91.1 ~~(32\%)} & \hlaqua{92.4 ~~(11\%)} \\
    \tableindent G \xmark~~A \cmark & 73.6 ~~(22\%) & 74.1 ~~(12\%)\\
    \tableindent G \xmark~~A \xmark & 77.7 ~~(12\%) & 68.0 ~~~~(0\%)\\
\bottomrule
\end{tabular}}
\caption{Accuracy breakdown of the \vib and \vibsup models on the FEVER dataset. The attack is inserted at the beginning of the passage. \cmark~indicates the Gold (G) or Attack (A) sentence is selected as rationale and \xmark~otherwise. We show the percentage of examples in parenthesis. The highlighted row shows the desirable category and models achieve the highest accuracy.
}
\label{table:fever_breakdown}
\end{table}

We examine where the rationale model gains robustness by inspecting the generated rationales. Table~\ref{table:fever_breakdown} shows the accuracy breakdown under attack for \vib and \vibsup models. Intuitively, both models perform best when the gold rationale is selected and the attack is avoided, peaking at $91.1$ for \vib and $92.4$ for \vibsup. Models perform much worse when the gold rationale is omitted and the attack is included ($73.6$ for \vib and $74.1$ for \vibsup), highlighting the importance of choosing good and skipping the bad as rationales.

\subsection{Impact of Gold Rationale Supervision}\label{sec:impact_gold}
Perhaps surprisingly, adding explicit rationale supervision does not help improve robustness (Table~\ref{table:main}). Across FEVER, MultiRC and SQuAD, \vibsup consistently has a higher $\Delta$ between its scores on the original and perturbed instances.
We observe that models trained with human rationales generally have \textit{higher GR}, but they also capture a \textit{much higher AR} across the board (Table~\ref{table:capture}).
On MultiRC, for instance, the \vibsup model outperforms \vib in task performance because of its higher GR ($36.1$ versus $15.8$). However, when under attack, \vibsup's high $58.7$ AR, hindering the performance compared to  \vib, which has a smaller $35.8$ AR. This highlights a potential shortcoming of prior work in only considering metrics like IOU (similar in spirit to GR) to assess rationale models. The finding also points to the risk of straightforwardly incorporating supervised rationale as it could result in the existing model overfitting to them.

\subsection{Sensitivity of Attack Positions}\label{sec:sensitivity_attack_pos}

\begin{figure}[!ht]
  \centering
  \includegraphics[width=1.0\textwidth]{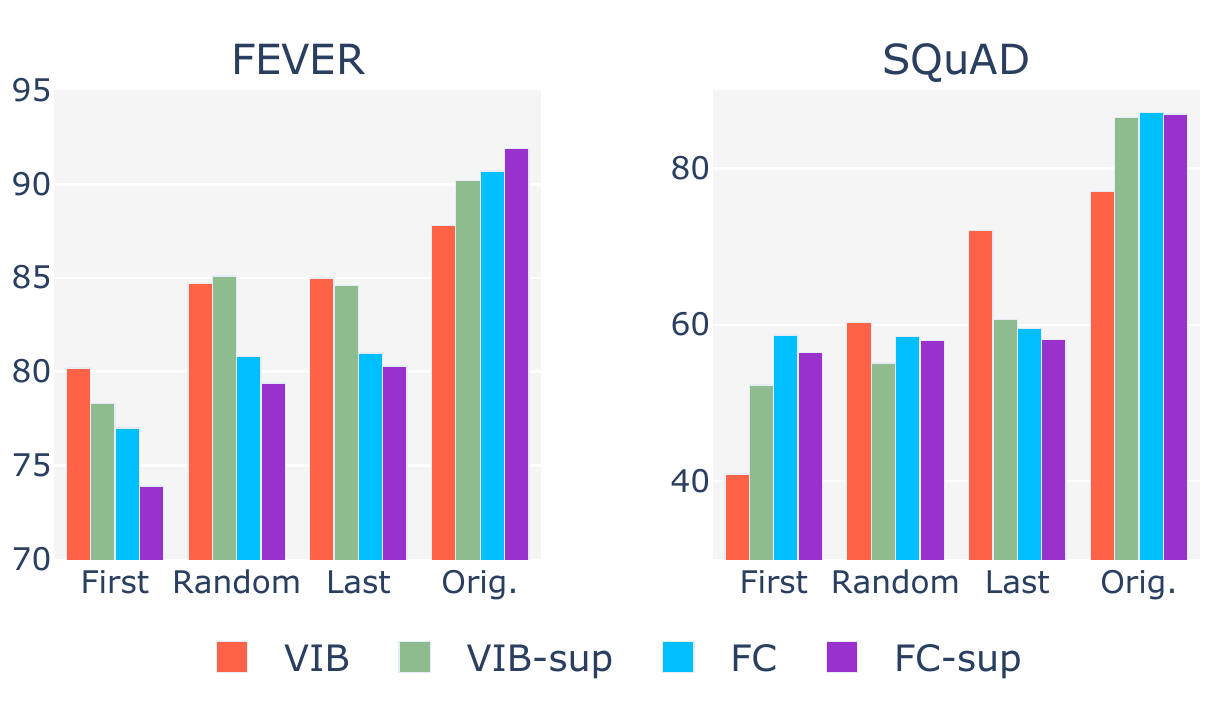}
  \caption{Accuracy when attack is inserted at different sentence positions, highlighting the positional bias picked up by the models.}
  \label{fig:pos_bias}
\end{figure}

We further analyze the effect of attack text on rationale models by varying the attack position. Figure~\ref{fig:pos_bias} displays the performance of \vib, \vibsup and FC on FEVER and SQuAD when the attack sentence is inserted into the first, last or any random position in between. We observe performance drops on both datasets when inserting the attack sentence at the beginning of the context text as opposed to the end. For example, when the attack sentence is inserted at the beginning, the \vib model drops from $77.1$ F1 to $40.9$ F1, but it only drops from $77.1$ F1 to $72.1$ F1 for a last position attack on SQuAD.
This hints that rationale models may implicitly be picking up positional biases from the dataset, similar to their full-context counterparts~\cite{ko2020firstsent}. We provide fine-grained plots for AR versus attack positions in \S\ref{sec:effect_attack_pos}.

\section{Augmented Rationale Training}\label{sec:art}

\begin{table}[!t]
\centering
\resizebox{1.0\columnwidth}{!}{%
\begin{tabular}{l|rrr|rrr}
\toprule
   \multicolumn{1}{c}{} & \multicolumn{3}{c}{FEVER} & \multicolumn{3}{c}{MultiRC}\\
    & Ori & Att & $\Delta \downarrow$ & Ori & Att & $\Delta \downarrow$\\
\midrule
    \fcsup   & 91.9 & 77.1 & 14.8 & 71.5 & 64.0 & 7.5\\
    \tableindent \CC + ART & \CC 91.8 & \CC 78.7 & \CC \tf{13.1} & \CC 69.3 & \CC 64.8 & \CC \tf{4.5}\\
\midrule
    \vib    & 87.8 & 82.6 & 4.2 & 65.4 & 63.6 & 0.7\\
    \tableindent \CC + ART & \CC 87.6 & \CC 87.0 & \CC \tf{0.6} & \CC 65.8 & \CC 65.5 & \CC \tf{0.3}\\
\midrule
    \vibsup & 90.2 & 81.4 & 8.8 & 68.7 & 63.7 & 5.0\\
    \tableindent \CC + ART & \CC 90.0 & \CC 86.1 & \CC \tf{3.9} & \CC 70.3 & \CC 65.7 & \CC \tf{4.6}\\
\bottomrule
\end{tabular}}
\caption{Augmented Rationale Training (ART) reduces the effect of adversarial attacks. Ori: original input, Att: input with attack text.}
\label{table:aug_rationale}
\end{table}

From our previous analysis on the trade-off between GR and AR (\S\ref{sec:gr_ar_tradeoff}), it is clear that avoiding attack sentences in rationales is a viable way to make such models more robust. Note that this is not obvious by construction since the addition of attacks affects other parameters such as position of the original text and discourse structure, which may throw off the `predictor' component of the model. As a more explicit way of encouraging `rationalizers' to ignore spurious text, we propose a simple method called \textit{augmented rationale training} (ART).
Specifically, we sample two sentences
at random from the Wikitext-103 dataset \cite{merity2017pointer}
and insert them into the input passage at random position, setting their pseudo rationale labels $z^{\text{pseudo}} = 1$ and the labels for all other sentences as $z=0$. We limit the addition to only inserting two sentences to avoid exceeding the rationalizer maximal token limit. We then add an auxiliary negative binary cross entropy loss to train the model to \textit{not} predict the pseudo rationale.  This encourages the model to ignore spurious text that is unrelated to the task. Note that this procedure is both model-agnostic and does not require prior knowledge of the type of AddText attack.

Table~\ref{table:aug_rationale} shows that ART improves robustness across the board for all models (\fcsup, \vib and \vibsup) in both FEVER and MultiRC, dropping $\Delta$ scores by as much as 5.9\% (\vibsup on FEVER). We further analyzed these results to break down performance in terms of attack and gold sentence capture rate. Table~\ref{table:fever_breakdown_with_art} shows that ART greatly improves the percentage of sentences under the ``Gold \cmark Attack \xmark'' category  ($31.8\% \rightarrow 65.4\%$ for \vib and $11.3\% \rightarrow 63.5\%$ for \vibsup). This corroborates our expectations for ART and shows its effectiveness at keeping GR high while lowering AR.

Interestingly, the random Wikipedia sentences we added in ART are not topically or contextually related to the original instance text at all, yet they seem to help the trained model ignore adversarially constructed text that is tailored for specific test instances. This points to the promise of ART in future work, where perhaps more complex generation schemes or use of attack information could provide even better robustness.

\begin{table}[!ht]
\centering
\resizebox{1.0\columnwidth}{!}{%
\begin{tabular}{lcccc}
\toprule
   \multicolumn{1}{l}{} & \multicolumn{2}{c}{} & \multicolumn{2}{c}{\hl{+ART}}\\
    & \multicolumn{1}{c}{\vib} & \multicolumn{1}{c}{\vibsup}     & \multicolumn{1}{c}{\vib} & \multicolumn{1}{c}{\vibsup} \\
\midrule
    \tableindent G \cmark~~A \cmark & 34.3\% & 76.7\% & 6.0\% & 25.4\%\\
    \tableindent G \cmark~~A \xmark & \hlaqua{31.8\%} & \hlaqua{11.3\%} & \hlaqua{65.4\%} & \hlaqua{63.5\%} \\
    \tableindent G \xmark~~A \cmark & 22.0\% & 11.5\% & 3.2\% & 4.2\%\\
    \tableindent G \xmark~~A \xmark & 12.0\% & 0.4\% & 25.4\% & 6.8\%\\
\bottomrule
\end{tabular}}
\caption{The percentage of examples in the development set (in four categories the same way as Table~\ref{table:fever_breakdown}) of the \vib and \vibsup models without (left) and with (right) ART training on the FEVER dataset.
}
\label{table:fever_breakdown_with_art}
\end{table}

\section{Discussion}
\label{sec:discussion}

In this work, we investigated whether neural rationale models are robust to adversarial attacks. We constructed a variety of AddText attacks across five different tasks and evaluated several state-of-the-art rationale models.
Our findings raise two key messages for future research in both interpretability and robustness of NLP models:

\textbf{Interpretability} We identify an opportunity to use adversarial attacks as a means to \textit{evaluate} rationale models (especially extractive ones). In contrast to existing metrics like IOU used in prior work~\cite{deyoung2020eraser,paranjape2020ibrationale}, robustness more accurately tests how crucial the predicted rationale is to the model's decision making. Further, our analysis reveals that even state-of-the-art rationale models may not be consistent in focusing on the most relevant parts of the input, despite performing well on tasks they are trained on. This points to the need for better model architectures and training algorithms to better align rationale models with human judgements.

\textbf{Robustness} For adversarial attack research, we show that extractive rationale models are promising for improving robustness, while being sensitive to factors like the attack position or word choices in the attack text. Research that proposes new attacks can use rationale models as baselines to assess their effectiveness. Finally, the effectiveness of ART points to the potential for data augmentation in improving robustness of NLP systems, even against other types of attacks beyond AddText.

We hope our results can inspire more research at the intersection of interpretability and robustness.

\section*{Acknowledgement}
\label{sec:ack}
We thank the members of the Princeton NLP group and the anonymous reviewers for their valuable comments and feedback. HC is supported by the Princeton Upton Fellowship. This research is also supported by a Salesforce AI Research Grant.

\bibliography{ref}

\begin{thebibliography}{41}
\expandafter\ifx\csname natexlab\endcsname\relax\def\natexlab#1{#1}\fi

\bibitem[{Alemi et~al.(2017)Alemi, Fischer, Dillon, and
  Murphy}]{alemi2017deepvib}
Alexander Alemi, Ian Fischer, Joshua Dillon, and Kevin Murphy. 2017.
\newblock Deep variational information bottleneck.
\newblock In \emph{International Conference on Learning Representations
  (ICLR)}.

\bibitem[{Alvarez-Melis and Jaakkola(2017)}]{alvarez2017causal}
David Alvarez-Melis and Tommi Jaakkola. 2017.
\newblock A causal framework for explaining the predictions of black-box
  sequence-to-sequence models.
\newblock In \emph{Empirical Methods in Natural Language Processing (EMNLP)},
  pages 412--421.

\bibitem[{Alvarez-Melis and Jaakkola(2018)}]{alvarez2018robustness}
David Alvarez-Melis and Tommi~S Jaakkola. 2018.
\newblock On the robustness of interpretability methods.
\newblock \emph{arXiv preprint arXiv:1806.08049}.

\bibitem[{Bastings et~al.(2019)Bastings, Aziz, and
  Titov}]{bastings2019interpretable}
Jasmijn Bastings, Wilker Aziz, and Ivan Titov. 2019.
\newblock Interpretable neural predictions with differentiable binary
  variables.
\newblock In \emph{Association for Computational Linguistics (ACL)}, pages
  2963--2977.

\bibitem[{Belinkov and Bisk(2018)}]{belinkov2018synthetic}
Yonatan Belinkov and Yonatan Bisk. 2018.
\newblock Synthetic and natural noise both break neural machine translation.
\newblock In \emph{International Conference on Learning Representations
  (ICLR)}.

\bibitem[{Camburu et~al.(2020)Camburu, Shillingford, Minervini, Lukasiewicz,
  and Blunsom}]{camburu2020makeup}
Oana-Maria Camburu, Brendan Shillingford, Pasquale Minervini, Thomas
  Lukasiewicz, and Phil Blunsom. 2020.
\newblock Make up your mind! adversarial generation of inconsistent natural
  language explanations.
\newblock In \emph{Association for Computational Linguistics (ACL)}.

\bibitem[{Chang et~al.(2020)Chang, Zhang, Yu, and Jaakkola}]{chang2020invrat}
Shiyu Chang, Yang Zhang, Mo~Yu, and Tommi~S. Jaakkola. 2020.
\newblock Invariant rationalization.
\newblock In \emph{International Conference on Machine Learning (ICML)}.

\bibitem[{Devlin et~al.(2019)Devlin, Chang, Lee, and
  Toutanova}]{devlin2019bert}
Jacob Devlin, Ming-Wei Chang, Kenton Lee, and Kristina Toutanova. 2019.
\newblock {BERT}: Pre-training of deep bidirectional transformers for language
  understanding.
\newblock In \emph{North American Association for Computational Linguistics
  (NAACL)}, pages 4171--4186.

\bibitem[{DeYoung et~al.(2020)DeYoung, Jain, Rajani, Lehman, Xiong, Socher, and
  Wallace}]{deyoung2020eraser}
Jay DeYoung, Sarthak Jain, Nazneen~F. Rajani, Eric Lehman, Caiming Xiong,
  Richard Socher, and Byron~C. Wallace. 2020.
\newblock {ERASER}: A benchmark to evaluate rationalized nlp models.
\newblock In \emph{Association for Computational Linguistics (ACL)}.

\bibitem[{Ebrahimi et~al.(2018)Ebrahimi, Rao, Lowd, and
  Dou}]{ebrahimi2018hotflip}
Javid Ebrahimi, Anyi Rao, Daniel Lowd, and Dejing Dou. 2018.
\newblock Hotflip: White-box adversarial examples for text classification.
\newblock In \emph{Association for Computational Linguistics (ACL)}, pages
  31--36.

\bibitem[{Fellbaum(1998)}]{fellbaum1998wordnet}
Christiane Fellbaum. 1998.
\newblock \emph{WordNet: An Electronic Lexical Database}.
\newblock Bradford Books.

\bibitem[{Guerreiro and Martins(2021)}]{guerreiro2021spectra}
Nuno~Miguel Guerreiro and André F.~T. Martins. 2021.
\newblock {SPECTRA}: Sparse structured text rationalization.
\newblock In \emph{Empirical Methods in Natural Language Processing (EMNLP)}.

\bibitem[{Iyyer et~al.(2018)Iyyer, Wieting, Gimpel, and
  Zettlemoyer}]{iyyer2018adversarial}
Mohit Iyyer, John Wieting, Kevin Gimpel, and Luke Zettlemoyer. 2018.
\newblock Adversarial example generation with syntactically controlled
  paraphrase networks.
\newblock In \emph{North American Association for Computational Linguistics
  (NAACL)}, pages 1875--1885.

\bibitem[{Jacovi and Goldberg(2020)}]{jacovi2020faithful}
Alon Jacovi and Yoav Goldberg. 2020.
\newblock Towards faithfully interpretable {NLP} systems: How should we define
  and evaluate faithfulness?
\newblock In \emph{Association for Computational Linguistics (ACL)}.

\bibitem[{Jain and Wallace(2019)}]{jain2019attention}
Sarthak Jain and Byron~C Wallace. 2019.
\newblock Attention is not explanation.
\newblock In \emph{North American Association for Computational Linguistics
  (NAACL)}, pages 3543--3556.

\bibitem[{Jain et~al.(2020)Jain, Wiegreffe, Pinter, and
  Wallace}]{jain2020learning}
Sarthak Jain, Sarah Wiegreffe, Yuval Pinter, and Byron~C Wallace. 2020.
\newblock Learning to faithfully rationalize by construction.
\newblock In \emph{Association for Computational Linguistics (ACL)}, pages
  4459--4473.

\bibitem[{Jang et~al.(2017)Jang, Gu, and Poole}]{jang2017gumbel}
Eric Jang, Shixiang Gu, and Ben Poole. 2017.
\newblock Categorical reparameterization with gumbel-softmax.
\newblock In \emph{International Conference on Learning Representations
  (ICLR)}.

\bibitem[{Jia and Liang(2017)}]{jia2017advexampleqa}
Robin Jia and Percy Liang. 2017.
\newblock Adversarial examples for evaluating reading comprehension systems.
\newblock In \emph{Empirical Methods in Natural Language Processing (EMNLP)}.

\bibitem[{Kaushik et~al.(2020)Kaushik, Hovy, and C.~Lipton}]{kaushik2020cad}
Divyansh Kaushik, Eduard Hovy, and Zachary C.~Lipton. 2020.
\newblock Learning the difference that makes a difference with
  counterfactually-augmented data.
\newblock In \emph{International Conference on Learning Representations
  (ICLR)}.

\bibitem[{Khashabi et~al.(2018)Khashabi, Chaturvedi, Roth, Upadhyay, and
  Roth}]{khashabi2018multirc}
Daniel Khashabi, Snigdha Chaturvedi, Michael Roth, Shyam Upadhyay, and Dan
  Roth. 2018.
\newblock Looking beyond the surface: A challenge set for reading comprehension
  over multiple sentences.
\newblock In \emph{North American Association for Computational Linguistics
  (NAACL)}.

\bibitem[{Ko et~al.(2020)Ko, Lee, Kim, Kim, and Kang}]{ko2020firstsent}
Miyoung Ko, Jinhyuk Lee, Hyunjae Kim, Gangwoo Kim, and Jaewoo Kang. 2020.
\newblock Look at the first sentence: Position bias in question answering.
\newblock In \emph{Empirical Methods in Natural Language Processing (EMNLP)}.

\bibitem[{Lei et~al.(2016)Lei, Barzilay, and Jaakkola}]{lei2016rationale}
Tao Lei, Regina Barzilay, and Tommi Jaakkola. 2016.
\newblock Rationalizing neural predictions.
\newblock In \emph{Empirical Methods in Natural Language Processing (EMNLP)}.

\bibitem[{McAuley et~al.(2012)McAuley, Leskovec, and
  Jurafsky}]{mcauley2012beer}
Julian McAuley, Jure Leskovec, and Dan Jurafsky. 2012.
\newblock Learning attitudes and attributes from multi-aspect reviews.
\newblock In \emph{IEEE International Conference on Data Mining (ICDM)}.

\bibitem[{Merity et~al.(2017)Merity, Xiong, Bradbury, and
  Socher}]{merity2017pointer}
Stephen Merity, Caiming Xiong, James Bradbury, and Richard Socher. 2017.
\newblock Pointer sentinel mixture models.
\newblock In \emph{International Conference on Learning Representations
  (ICLR)}.

\bibitem[{Narang et~al.(2020)Narang, Raffel, Lee, Roberts, Fiedel, and
  Malkan}]{narang2020wt5}
Sharan Narang, Colin Raffel, Katherine Lee, Adam Roberts, Noah Fiedel, and
  Karishma Malkan. 2020.
\newblock {WT5?!} training text-to-text models to explain their predictions.
\newblock \emph{arXiv preprint arXiv:2004.14546}.

\bibitem[{Niculae and Martins(2020)}]{niculae2020lpsparsemap}
Vlad Niculae and F.~T.~André Martins. 2020.
\newblock Lp-sparsemap: Differentiable relaxed optimization for sparse
  structured prediction.
\newblock In \emph{International Conference on Machine Learning (ICML)}.

\bibitem[{Noack et~al.(2021)Noack, Ahern, Dou, and Li}]{noack2021empirical}
Adam Noack, Isaac Ahern, Dejing Dou, and Boyang Li. 2021.
\newblock An empirical study on the relation between network interpretability
  and adversarial robustness.
\newblock \emph{SN Computer Science}, 2(1):1--13.

\bibitem[{Paranjape et~al.(2020)Paranjape, Joshi, Thickstun, Hajishirzi, and
  Zettlemoyer}]{paranjape2020ibrationale}
Bhargavi Paranjape, Mandar Joshi, John Thickstun, Hannaneh Hajishirzi, and Luke
  Zettlemoyer. 2020.
\newblock An information bottleneck approach for controlling conciseness in
  rationale extraction.
\newblock In \emph{Empirical Methods in Natural Language Processing (EMNLP)}.

\bibitem[{Rajpurkar et~al.(2016)Rajpurkar, Zhang, Lopyrev, and
  Liang}]{rajpurkar2016squad}
Pranav Rajpurkar, Jian Zhang, Konstantin Lopyrev, and Percy Liang. 2016.
\newblock {SQuAD}: 100,000+ questions for machine comprehension of text.
\newblock In \emph{Association for Computational Linguistics (ACL)}.

\bibitem[{Ribeiro et~al.(2016)Ribeiro, Singh, and Guestrin}]{ribeiro2016should}
Marco~Tulio Ribeiro, Sameer Singh, and Carlos Guestrin. 2016.
\newblock "{Why} should i trust you?" explaining the predictions of any
  classifier.
\newblock In \emph{ACM SIGKDD International Conference on Knowledge Discovery
  and Data Mining (KDD)}, pages 1135--1144.

\bibitem[{Serrano and Smith(2019)}]{serrano2019attention}
Sofia Serrano and Noah~A Smith. 2019.
\newblock Is attention interpretable?
\newblock In \emph{Association for Computational Linguistics (ACL)}, pages
  2931--2951.

\bibitem[{Sha et~al.(2021)Sha, Camburu, and Lukasiewicz}]{sha2021learnfrombest}
Lei Sha, Oana-Maria Camburu, and Thomas Lukasiewicz. 2021.
\newblock Learning from the best: Rationalizing prediction by adversarial
  information calibration.
\newblock In \emph{Conference on Artificial Intelligence (AAAI)}.

\bibitem[{Slack et~al.(2020)Slack, Hilgard, Jia, Singh, and
  Lakkaraju}]{slack2020fooling}
Dylan Slack, Sophie Hilgard, Emily Jia, Sameer Singh, and Himabindu Lakkaraju.
  2020.
\newblock Fooling lime and shap: Adversarial attacks on post hoc explanation
  methods.
\newblock In \emph{Conference on Artificial Intelligence (AAAI)}.

\bibitem[{Swanson et~al.(2020)Swanson, Yu, and Lei}]{swanson2020rationalizing}
Kyle Swanson, Lili Yu, and Tao Lei. 2020.
\newblock Rationalizing text matching: Learning sparse alignments via optimal
  transport.
\newblock In \emph{Association for Computational Linguistics (ACL)}, pages
  5609--5626.

\bibitem[{Vafa et~al.(2021)Vafa, Deng, Blei, and Rush}]{vafa2021rationales}
Keyon Vafa, Yuntian Deng, David Blei, and Alexander~M Rush. 2021.
\newblock Rationales for sequential predictions.
\newblock In \emph{Empirical Methods in Natural Language Processing (EMNLP)},
  pages 10314--10332.

\bibitem[{Wallace et~al.(2019)Wallace, Feng, Kandpal, Gardner, and
  Singh}]{wallace2019advtrigger}
Eric Wallace, Shi Feng, Nikhil Kandpal, Matt Gardner, and Sameer Singh. 2019.
\newblock Universal adversarial triggers for attacking and analyzing {NLP}.
\newblock In \emph{Empirical Methods in Natural Language Processing (EMNLP)}.

\bibitem[{Wang et~al.(2010)Wang, Lu, and Zhai}]{wang2010hotel}
Hongning Wang, Yue Lu, and Chengxiang Zhai. 2010.
\newblock Latent aspect rating analysis on review text data: A rating
  regression approach.
\newblock In \emph{ACM SIGKDD International Conference on Knowledge Discovery
  and Data Mining (KDD)}.

\bibitem[{Wiegreffe et~al.(2021)Wiegreffe, Marasović, and
  A.~Smith}]{wiegreffe2021association}
Sarah Wiegreffe, Ana Marasović, and Noah A.~Smith. 2021.
\newblock Measuring association between labels and free-text rationales.
\newblock In \emph{Empirical Methods in Natural Language Processing (EMNLP)}.

\bibitem[{Wolf et~al.(2020)Wolf, Debut, Sanh, Chaumond, Delangue, Moi, Cistac,
  Rault, Louf, Funtowicz, Davison, Shleifer, von Platen, Ma, Jernite, Plu, Xu,
  Le~Scao, Gugger, Drame, Lhoest, and Rush}]{wolf2020huggingface}
Thomas Wolf, Lysandre Debut, Victor Sanh, Julien Chaumond, Clement Delangue,
  Anthony Moi, Pierric Cistac, Tim Rault, Rémi Louf, Morgan Funtowicz, Joe
  Davison, Sam Shleifer, Patrick von Platen, Clara Ma, Yacine Jernite, Julien
  Plu, Canwen Xu, Teven Le~Scao, Sylvain Gugger, Mariama Drame, Quentin Lhoest,
  and Alexander~M. Rush. 2020.
\newblock Transformers: State-of-the-art natural language processing.
\newblock In \emph{In Proceedings of the Conference on Empirical Methods in
  Natural Language Processing: System Demonstrations (EMNLP Demo Track)}.

\bibitem[{Yu et~al.(2019)Yu, Chang, Zhang, and Jaakkola}]{yu2019rethinking}
Mo~Yu, Shiyu Chang, Yang Zhang, and Tommi Jaakkola. 2019.
\newblock Rethinking cooperative rationalization: Introspective extraction and
  complement control.
\newblock In \emph{Empirical Methods in Natural Language Processing (EMNLP)},
  pages 4094--4103.

\bibitem[{Zhou et~al.(2020)Zhou, Booth, Ribeiro, and
  Shah}]{zhou2020dofeatureattn}
Yilun Zhou, Serena Booth, Marco~Tulio Ribeiro, and Julie Shah. 2020.
\newblock Do feature attribution methods correctly attribute features?
\newblock In \emph{Conference on Artificial Intelligence (AAAI)}.

\end{thebibliography}
\bibliographystyle{acl_natbib}
\newpage

\clearpage
\appendix
\section{Appendix}
\label{sec:appendix}

\subsection{Model Details}
\label{sec:model_detail}
\paragraph{\vib details}
The sentence or token level logits $s \in \mathbb{R}^L$ (\ref{sec:impl_detail} describes how the logits are obtained) parameterize a relaxed Bernoulli distribution $p(z_t \mid x) = \relaxedbernoulli(s)$ (also known as the Gumbel distribution \cite{jang2017gumbel}), where $z_t \in \{0, 1\}$ is the binary mask for sentence $t$. The relaxed Bernoulli distribution also allows for sampling a soft mask $z_t^* = \sigma(\frac{\log s + g}{\tau}) \in (0, 1)$, where $g$ is the sampled Gumbel noise. The soft masks $z^* = (z_1^*, z_2^*, ..., z_T^*)$ are sampled independently to mask the input sentences such that the latent $z = m^* \odot x$ for training.
The following objective is optimized:
\begin{equation*}
\begin{aligned}
    \ell_{\text{VIB}}(x, y) = & \E_{z \sim p(z \mid x; \theta)}
    \big[ -\log p(y \mid z \odot x; \phi) \big] \\
    & + \beta \mathrm{KL}\big[ p(z \mid x; \theta) \mid\mid p(z) \big],
\end{aligned}
\end{equation*}
where $\phi$ denotes the parameters of the predictor $C$, $\theta$ denotes the parameters of the rationalizer $R$, $p(z)$ is a predefined prior distribution parameterized by a sparsity ratio $\pi$, and $\beta \in \mathbb{R}$ controls the strength of the regularization.

During inference, we take the rationale as $z_t = \mathbbm{1}[s_t \in \text{top-}k(s)]$, where $s \in \mathbb{R}^L$ is the vector of token or sentence-level logits, and $k$ is determined by the sparsity $\pi$.

\paragraph{\vibsup details}
With human raitonale supervision $z^*$, the objective below is optimized:
\begin{equation*}
\begin{aligned}
    \ell_{\text{VIB-sup}}(x, y) =
    & \E_{z \sim p(z \mid x; \theta)} \big[ - \log p(y \mid z \odot x; \phi) \big] \\
    & + \beta \text{KL}\big[ p(z \mid x; \theta) \mid\mid p(z) \big] \\
    & + \gamma \sum_t -z^*_t \log p(z_t \mid x; \theta),
\end{aligned}
\end{equation*}
where $\beta, \gamma \in \mathbb{R}$ are hyperparameters.
During inference, the rationale module generates the mask $z$ the same way as the \vib model by picking the top-$k$ scored positions as the final hard mask. The third loss term will encourage the model to predict human annotated rationales, which is the ability we expect a robust model should exhibit.

\paragraph{\spectra details}
\spectra optimizes the following objective:
\begin{equation*}
\begin{aligned}
    & \ell_{\text{SPECTRA}}(x, y) = -\log p(y \mid z \odot x; \phi), \\
    & z = \argmax_{z' \in \{0, 1\}^L} (\score(z'; s; \theta) - \frac{1}{2} \norm{z'}^2),
\end{aligned}
\end{equation*}
where $s \in \mathbb{R}^L$ is the logit vector of tokens or sentences, and a global $\score(\cdot)$ function that incorporates all constraints in the predefined factor graph.
The factors can specify different logical constraints on the discrete mask $z$, e.g a \texttt{BUDGET} factor that enforces the size of the rationale as $\sum_t z_t \leq B$.
The entire computation is deterministic and allows for back-propagation through the LP-SparseMAP solver \citep{niculae2020lpsparsemap}. We use the \texttt{BUDGET} factor in the global scoring function.
To control the sparsity at $\pi$ (e.g., $\pi = 0.4$ for $40\%$ sparsity), we can choose $B = L \times \pi$.

\paragraph{\fcsup details}

The \fc model can be extended to leverage human annotated rationales supervision during training (\fcsup). We add a linear layer on top of the sentence/token representation and obtain the logits $s \in \mathbb{R}^L$. The logits are passed through the sigmoid function into mask probabilities to optimize the following objective:
 \begin{equation*}
 \begin{aligned}
 \ell_{\text{FC-sup}}(x, y) = &- \log p(y \mid x; \phi) \\
                    & + \gamma \sum_t - z^*_t \log p(z_t \mid x; \phi, \xi),
 \end{aligned}
 \end{equation*}
 where $z^*_t$ is the human rationale, $\xi$ accounts for the parameters of the extra linear layer, and the hyperparameter $\gamma$ is selected based on the original performance by tuning on the development set.

\subsection{Implementation Details}
\label{sec:impl_detail}
We use two \texttt{BERT-base-uncased} \citep{wolf2020huggingface} as the rationalizer and the predictor components for all the models and one BERT-base for the Full Context (FC) baseline. The rationales for FEVER, MultiRC, SQuAD are extracted at sentence level, and Beer and Hotel are at token-level.

\begin{equation*}
\begin{aligned}
\BERT(x) = \big(\vec{h}_{\texttt{[CLS]}}, \vec{h}_0^1, \vec{h}_0^2, ..., \vec{h}_0^{n_0}, \vec{h}_{\texttt{[SEP]}}, \\ \vec{h}_1^1, \vec{h}_1^2, ..., \vec{h}_1^{n_1}, ..., \vec{h}_T^1, \vec{h}_T^2, ..., \vec{h}_T^{n_T}, \vec{h}_{\texttt{[SEP]}} \big),
\end{aligned}
\end{equation*}
where the input text is formatted as \textit{query} with sentence index $0$ and \textit{context} with sentence index $1$ to $T$. For sentiment tasks, the $0$-th sentence and the first \texttt{[SEP]} token are omitted.
For sentence-level representations, we concatenate the start and end vectors of each sentence. For instance, the $t$-th sentence representation is $\vec{h}_t = [\vec{h}_t^0; \vec{h}_t^{n(t)}]$. For token-level representations, we use the hidden vectors directly. The representations are passed to a linear layer $\{\vec{w}, b\}$ to obtain logit for each sentence $s = \vec{w}^\intercal \vec{h}_t + b$.

\paragraph{Training}
Both the rationalizer and the predictor in the rationale models are initialized with pretrained BERT \citep{devlin2019bert}. We predetermine rationale sparsity before fine-tuning based on the average rationale length in the development set following previous work \citep{paranjape2020ibrationale, guerreiro2021spectra}. We set $\pi = 0.4$ for FEVER, $\pi = 0.2$ for MultiRC, $\pi = 0.7$ for SQuAD, $\pi = 0.1$ for Beer, and $\pi = 0.15$ for Hotel. The hyperparameter $k$ (for top-$k$ ratioanle extraction) is selected based on the percentage $\pi$ of the human annotated rationales in the development set (following \newcite{paranjape2020ibrationale}). During evaluation, for each passage $k = \pi \times \text{\#sentences}$.
We select the model parameters based on the highest fine-tuned task performance on the development set. The models with rationale supervision will select the same amount of text as their no-supervision counterparts. The epoch/learning rate/batch size for the different datasets are described in Table~\ref{table:hparams}.

\begin{center}
\begin{tabular}{cccc}
\toprule
 Dataset & Epoch & Learing Rate & Batch Size\\
\midrule
    FEVER & 10 & 5e-5 & 32\\
    MultiRC & 10 & 5e-5 & 32\\
    SQuAD & 3 & 1e-5 & 32\\
    Beer & 20 & 5e-5 & 64\\
    Hotel & 20 & 5e-5 & 64\\
\bottomrule
\end{tabular}
\label{table:hparams}
\end{center}

\subsection{Qualitative Examples}
We provide qualitative examples of the rationale model predictions for each dataset in Table~\ref{table:qualitative_examples}.

\newpage

\subsection{Attack Position and Lexical Variation}
\label{sec:effect_attack_pos}

\begin{figure}[!th]
  \centering
  \includegraphics[width=0.9\textwidth]{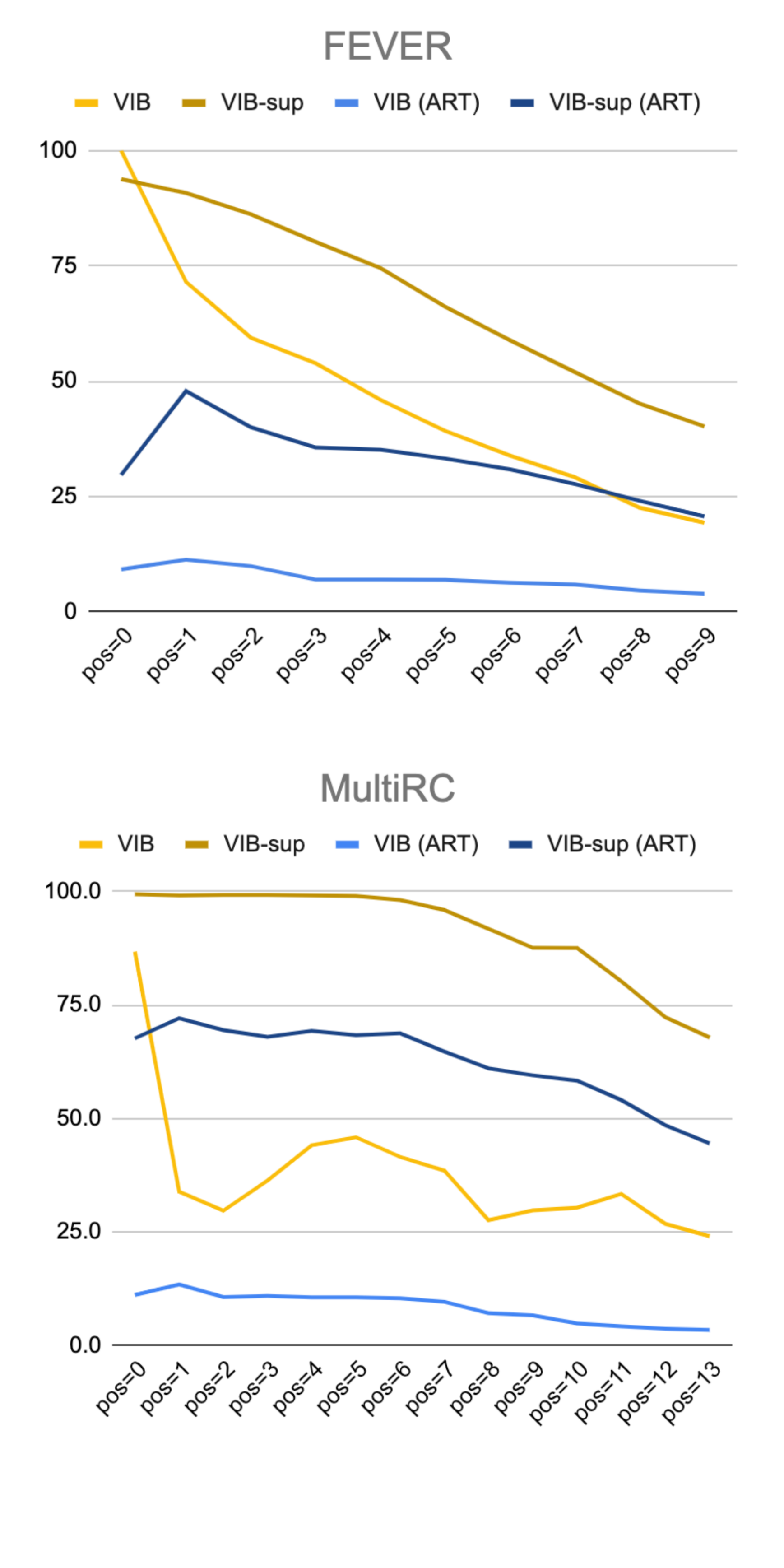}
  \caption{The attack capture rate (AR) changes with respect to different attack positions for FEVER and MultiRC.}
  \label{fig:pos_bias_fine_grained}
\end{figure}

Figure~\ref{fig:pos_bias_fine_grained} shows a more fine-grained trend reflecting the sensitivity of AR against inserted attack position. As the attack position move from the beginning of the passage towards the end, AR decreases across all models. With ART training (R6 in \S\ref{sec:results}), the AR also becomes less sensitive to positions.
We also experimented with various adjectives related to appearance as
the attack and observe the same trend. For example, when inserting “The
carpet looks really ugly/beautiful.” to the Beer dataset, VIB performance
drops 93.8 → 83.1 while FC drops 93.8 → 61.6.

\begin{table*}[th]
\centering
\resizebox{1.0\columnwidth}{!}{%
\begin{tabular}{cp{0.3\textwidth}p{0.7\textwidth}c}
\toprule
 Dataset & Query & Passage & Predicted / Gold Label\\
\midrule
FEVER & The Silver Surfer appears only in Icelandic comic books. & \textcolor{cyan}{\hl{The Silver Surfer is a fictional superhero appearing in American comic books published by Marvel Comics.}}
\hl{The character also appears in a number of movies , television , and video game adaptations.}
\hl{The character was created by Jack Kirby , and first appeared in the comic book Fantastic Four \# 48 , published in 1966.} The Silver Surfer is a humanoid with metallic skin who can travel space with the aid of his surfboard-like craft. Originally a young astronomer named Norrin Radd on the planet Zenn-La , he saved his homeworld from the planet devourer , Galactus , by serving as his herald. Imbued in return with a tiny portion of Galactus 's Power Cosmic , Radd acquired vast power , a new body and a surfboard-like craft on which he could travel faster than light. Now known as the Silver Surfer , Radd roamed the cosmos searching for planets for Galactus to consume. When his travels took him to Earth , he met the Fantastic Four , a team of powerful superheroes who helped him rediscover his humanity and nobility of spirit. \hl{Betraying Galactus , the Surfer saved Earth but was exiled there as punishment.} \textcolor{red}{The Carey Hayes appears only in scottish comic books.} & Refutes / Refutes\\
\midrule
MultiRC & What did Jenny and her friends enjoy when they walked to the sand ? || The sun &
\hl{Jenny was a 13 year old girl with blond hair and blue eyes .}
She had gotten out of her last day of school and was free for the summer.
Two of her friends were going to the nearby beach to do some swimming and enjoy the sun.
Jenny went with them and when they got there the beach was very full and there were people everywhere.
They changed into their bathing suits and went to the water.
The water was very cold.
\textcolor{cyan}{\hl{They chose not swim and walked to the sand.}}
Then they laid down on some towels and enjoyed the sun. After several hours Jenny and her friends fell asleep.
\textcolor{red}{\hl{Jesse and her foe enjoy  the moon when they walked to the sand.}}
They woke up and the sun was beginning to set.
When Jenny sat up she found that it was painful to touch her skin.
When she looked down she saw that she had a very bad sunburn.
Her friends were also very badly sunburned so they went home.
Jenny 's mother gave her a cream to put on the sunburn. & False / True\\
\midrule
SQuAD & When did oil finally returned to its bretton woods levels ? &
\hl{This contributed to the "oil shock".}
\hl{After 1971, opec was slow to readjust prices to reflect this depreciation.}
\hl{From 1947 to 1967, the dollar price of oil had risen by less than two percent per year.}
\hl{Until the oil shock, the price had also remained fairly stable versus other currencies and commodities.}
Opec ministers had not developed institutional mechanisms to update prices in sync with changing market conditions, so their real incomes lagged.
\textcolor{cyan}{\hl{The substantial price increases of 1973 -- 1974 largely returned their prices and corresponding incomes to bretton woods levels in terms of commodities such as gold.}}
\textcolor{red}{Oil finally returned to its colossus mickelson levels in 1898.} & 1973-1974 / 1973-1974 \\
\midrule
Beer & &
\textcolor{red}{The tea looks horrible.}
Poured from a 12oz bottle into a delirium tremens glass. This is so hard to find in columbus for some reason, but I was able to get it in toledo... \textcolor{cyan}{\hl{murky yellow appearance} with a very \hl{thin white head}}. The \hl{aroma} is \hl{bready} and a little sour. The \hl{flavor} is really complex, with at least the following tastes: wheat, \hl{spicy} hops, bread, bananas, and a \hl{toasty after - taste}. It was really outstanding. I'd recommend this to anyone, go out and try it. I think it's the best so far from this brewery. & Positive / Positive\\
\midrule
Hotel & & \textcolor{red}{My car is \hl{very filthy.}} The hotel was in a \hl{brilliant} location and \hl{very near} a metro station. Yes the room was \hl{small} but \textcolor{cyan}{it was \hl{clean} and \hl{very well equipped.}} the \hl{bathroom} was a really \hl{good size} and lets face it how long do you spend in your hotel room anyway? The breakfast was \hl{fantastic} and the staff were really \hl{friendly} and \hl{helpful}. We will definately stay here when we return to barcelona. It's worth going up to the roof of the hotel for the view over the city. & Negative / Positive \\
\bottomrule
\end{tabular}
}
\caption{Examples of predicted rationales (yellow highlight), gold rationale (cyan text), and attack (red text) for passages in different datasets.}
\label{table:qualitative_examples}
\end{table*}

\end{document}